\documentclass{article}

\usepackage{microtype}
\usepackage{graphicx}
\usepackage{cuted}
\usepackage{subcaption}
\usepackage{booktabs} 

\usepackage{hyperref}

\usepackage[preprint]{icml2027}

\usepackage{amsmath}
\usepackage{amssymb}
\usepackage{mathtools}
\usepackage{amsthm}
\usepackage{algorithm}
\usepackage{algorithmic}
\usepackage{xcolor}
\usepackage{bm}

\usepackage[capitalize,noabbrev]{cleveref}

\theoremstyle{plain}

\theoremstyle{definition}

\theoremstyle{remark}

\usepackage[textsize=tiny]{todonotes}

\newcommand{\Mat}{\mathcal{M}}
\newcommand{\Lat}{L}
\newcommand{\Frac}{F}
\newcommand{\Atom}{A}

\newcommand{\NumOfTypes}{N_{\text{types}}}

\newcommand{\Normal}{\mathcal{N}}
\newcommand{\WN}{\mathcal{W}\mathcal{N}}
\newcommand{\Cat}{\text{Cat}}
\newcommand{\Unif}{\mathcal{U}}
\newcommand{\Loss}{\mathcal{L}}
\newcommand{\Mean}{\mathbb{E}}
\newcommand{\KL}{\text{KL}}
\newcommand{\one}{\boldsymbol{1}}

\icmltitlerunning{DAMi: RL Fine-Tuning of the MiAD Crystal Diffusion Model}

\begin{document}

\twocolumn[
  \icmltitle{DAMi: RL Fine-Tuning of the MiAD Crystal Diffusion Model}



  \icmlsetsymbol{equal}{*}

  \begin{icmlauthorlist}
    \icmlauthor{Andrei Nazarov}{equal,hse}
    \icmlauthor{Maksim Nakhodnov}{equal,constructor}
    \icmlauthor{Nikita Morozov}{hse}
    \icmlauthor{Karina Kuchma}{msu}
    \icmlauthor{Nikita Kazeev}{nus}
    \icmlauthor{Andrey Ustyuzhanin}{constructor}
    \icmlauthor{Mikhail Lazarev}{hse}
    \icmlauthor{Dmitry Vetrov}{constructor}
    \icmlauthor{Andrey Okhotin}{hse,msu}
  \end{icmlauthorlist}

  \icmlaffiliation{hse}{HSE University}
  \icmlaffiliation{msu}{Moscow State University}
  \icmlaffiliation{constructor}{Constructor University of Bremen}
  \icmlaffiliation{nus}{National University of Singapore}

  \icmlcorrespondingauthor{Andrey Okhotin}{andrey.okhotin@gmail.com}

  \icmlkeywords{Machine Learning, ICML}

  \vskip 0.3in
]



\printAffiliationsAndNotice{\icmlEqualContribution}

\begin{abstract}
Crystal discovery requires a generator to produce stable structures while continuing to sample distinct candidates. We introduce DAMi, a reinforcement learning approach to fine-tuning the pretrained MiAD crystal diffusion model with a stability reward derived from CHGNet energies. We adapt diffusion policy optimization with KL regularization (DPOK) and relative trajectory balance (RTB) to MiAD by evaluating denoising transitions for the lattice, periodic fractional coordinates, and atom types. Because the fraction of stable, unique, and novel (SUN) crystals depends on the generation budget, we evaluate checkpoints at multiple finite budgets rather than a single draw count. RTB fine-tuned MIAD achieves the highest SUN rates in this comparison at 10,000 and 100,000 draws, reaching 15.55\% and 10.32\%, respectively, compared with 15.05\% and 7.70\% for Chemeleon2-GRPO.

\end{abstract}

\section{Introduction}

Automated materials discovery pipelines typically combine a proposal stage with downstream filtering, relaxation, and experimental or high-fidelity validation.
Generative models occupy the proposal stage. They sample candidate crystal structures that can be passed to cheaper surrogates and later to more expensive checks~\citep{zeni2025generative,jiao2023crystal,okhotin2025miad}.
Their role is to produce chemically plausible structures and to keep supplying candidates that remain useful after uniqueness and novelty filtering.

Most crystal generators are trained with supervised objectives that imitate structures from curated crystallographic datasets.
This strategy is limited by the size and coverage of collections of known stable materials. The generator can only learn what those datasets contain, and expanding them with DFT is expensive.
An alternative is to fine-tune a pretrained generator with reinforcement learning, using a machine-learned interatomic potential (MLIP) as a reward model that scores energies or hull distances~\citep{deng2023chgnet,park2025guiding,prakash2026reinforcement}.
Modern MLIPs are trained on far larger corpora of labeled structures and trajectories than the stable-material sets used for generator pretraining, including resources such as MPtrj~\citep{deng2023chgnet}, OMat24~\citep{barroso2024omat24}, and sAlex~\citep{barroso2024omat24,schmidt2024alexandria}.
This scale makes energy-based rewards a practical signal for steering generation toward lower-energy regions of chemical space.

However, RL also gives the generator freedom to exploit the reward.
Without sufficient regularization or diversity pressure, optimization can collapse onto a few high-reward modes. Short-horizon stability then improves while long-horizon discovery degrades~\citep{govindarajan2026crystune}.
Understanding this behavior requires metrics that track how the yield of stable, unique, and novel (SUN) crystals evolves with the generation budget. It also requires training choices that mitigate collapse while still improving stability.

We fine-tune the pretrained MiAD crystal diffusion model with two such methods, DPOK and RTB, using a CHGNet-based stability reward and a frozen MiAD reference.
We measure SUN rates at 2,000, 10,000, and 100,000 draws to expose the finite-budget trade-off between early gains and sustained discovery.

\paragraph{Contributions}
\begin{enumerate}
    \item We adapt DPOK and RTB to MiAD's multi-component diffusion with a CHGNet stability reward.
    \item We show that SUN@budget rankings change with the generation budget, so a checkpoint that looks best at a small draw count need not remain best at larger budgets.
    \item We show that selected MiAD-DPOK and MiAD-RTB checkpoints outperform Chemeleon2-GRPO at SUN@10k and SUN@100k, with MiAD-RTB best on both.
\end{enumerate}

\section{RL Finetuning of Diffusion Models}
\label{sec:rl-finetuning}

\paragraph{Diffusion models.}
Diffusion models learn to generate data by reversing a gradual corruption process.
Its forward and backward processes factorize as follows:
\begin{align}
    q(x_{0:T})
        &=q(x_0)\prod_{t=1}^{T}q(x_t\mid x_{t-1}), \notag\\
    p_\theta(x_{0:T})
        &=p(x_T)\prod_{t=1}^{T}p_\theta(x_{t-1}\mid x_t),
    \label{eq:diffusion-processes}
\end{align}
Generation starts from a tractable prior $p_T$ and follows the reverse transitions to obtain $x_0$.
Diffusion models can be formulated for both continuous and discrete variables, which makes them well suited to heterogeneous data.

\paragraph{RL interpretation.}
The reverse chain defines a sequential decision process with state $(x_t,t)$, action $x_{t-1}$, and policy $p_\theta(x_{t-1}\mid x_t)$.
A terminal reward $r(x_0)$ evaluates the generated sample.
Policy gradients use the log-probabilities of denoising trajectories $\tau=x_{0:T}$ to improve expected reward without differentiating through $r$.

The two methods used in this work, DPOK and RTB, share a trade-off between increasing expected reward and limiting deviation from the frozen pretrained model. For an output distribution $\pi$ over $x_0$ and a regularization strength $\beta>0$, this trade-off is expressed by
\begin{equation}
    J(\pi)=\Mean_{x_0\sim \pi}[r(x_0)]-\beta\KL(\pi\Vert p_{\mathrm{ref}}),
    \label{eq:reward-kl}
\end{equation}
where $p_{\mathrm{ref}}(x_0)$ is the pretrained output distribution. The maximizer over unrestricted output distributions is $\pi^\star(x_0)\propto p_{\mathrm{ref}}(x_0)\exp(r(x_0)/\beta)$, since $J(\pi)$ equals $-\beta\KL(\pi\Vert \pi^\star)$ plus a constant independent of $\pi$.

\paragraph{Stability reward.}
We use CHGNet~\citep{deng2023chgnet} to estimate the energy per atom of each crystal $x_0$ relative to the reference convex hull. Both methods use $r(x_0)=-\tanh((E-0.1)/0.2)$, where $E$ is this energy in eV/atom, and 0.1 eV/atom is crystal metastability threshold.

\subsection{DPOK}

Diffusion Policy Optimization with KL regularization (DPOK) uses a policy-gradient update with a tractable penalty on the denoising transitions. The expected sum of transition KL divergences upper-bounds the output KL in Eq.~\eqref{eq:reward-kl} \citep[Lemma~4.2]{fan2023dpok}. Following \citet[Eq.~9]{fan2023dpok}, the gradient estimate is
\begin{equation}
\begin{aligned}
    g_\theta=&\Mean_{\tau\sim p_\theta}\Bigl[
    -r(x_0)\sum_{t=1}^{T}
        \nabla_\theta\log p_\theta(x_{t-1}\mid x_t)\\
    &+\beta\sum_{t=1}^{T}\nabla_\theta
        \KL\!\left(p_\theta(x_{t-1}\mid x_t)\Vert p_{\mathrm{ref}}(x_{t-1}\mid x_t)\right)
    \Bigr].
\end{aligned}
\label{eq:dpok-gradient}
\end{equation}
The reward term increases the probability of denoising transitions along trajectories that yield higher terminal rewards.
The KL term penalizes deviations from the pretrained transition distributions.
The expectation is over trajectories sampled from the current policy.

\subsection{RTB}

Relative trajectory balance (RTB; \citealp{venkatraman2024amortizing}) targets $\pi^\star$ by reweighting the pretrained trajectory distribution by $\exp(r(x_0)/\beta)$. The squared residual for one trajectory and its training expectation are
\begin{align}
    \ell_{\mathrm{RTB}}(\tau;\theta)
        &=\left[\log\frac{Z_\theta p_\theta(\tau)}{p_{\mathrm{ref}}(\tau)}
          -\frac{r(x_0)}{\beta}\right]^2, \notag\\
    \Loss_{\mathrm{RTB}}(\theta)
        &=\Mean_{\tau\sim\mu}\!\left[\ell_{\mathrm{RTB}}(\tau;\theta)\right].
    \label{eq:rtb-loss}
\end{align}
Here $Z_\theta$ is a normalizing constant learned in log space. The sampling distribution $\mu$ may supply on-policy or off-policy trajectories. In either case the gradient is taken with the sampled trajectory held fixed \citep[Eq.~9 and Sec.~2.3]{venkatraman2024amortizing}.
With a shared initial distribution $p_T$, the trajectory log-ratio is a sum of transition log-ratios.
If the residual vanishes for every trajectory, the model satisfies
\begin{equation}
    Z_\theta p_\theta(\tau)
    =p_{\mathrm{ref}}(\tau)\exp\!\left(\frac{r(x_0)}{\beta}\right).
    \label{eq:rtb-optimum}
\end{equation}
Integrating out $x_{1:T}$ then gives $p_\theta(x_0)=\pi^\star(x_0)$, the maximizer of Eq.~\eqref{eq:reward-kl} \citep[Prop.~1]{venkatraman2024amortizing}.

Because MiAD models the three components of a crystal, applying DPOK and RTB requires computing the log-probabilities of the transition distributions of the backward processes for each component.
\section{MiAD}
\label{sec:miad}

MiAD is a joint diffusion model over crystal structures $\Mat = (\Lat, \Frac, \Atom)$ \citep{jiao2023crystal, zeni2025generative}.
The forward noising process factorizes across the lattice $\Lat$, fractional coordinates $\Frac$, and atom types $\Atom$:
\begin{align*}
    q(\Mat_t|\Mat_{t-1}) = q(\Lat_t|\Lat_{t-1})q(\Frac_t|\Frac_{t-1})q(\Atom_t|\Atom_{t-1}),
\end{align*}
where $\Mat_0 = (\Lat_0, \Frac_0, \Atom_0)$ is a clean crystal and $\Mat_t = (\Lat_t, \Frac_t, \Atom_t)$ is its noisy counterpart at step $t$.
The reverse predictors use the full crystal state $\Mat_t$.
Although their parameters are predicted jointly, the three components have different transition distributions.

\subsection{Lattice}
Since $\Lat_0 \in \mathbb{R}^{3\times3}$, the lattice component uses a standard DDPM with Gaussian transitions \citep{ho2020denoising}:
\begin{align*}
    q(\Lat_t|\Lat_{t-1}) & = \Normal\left(\Lat_t|\sqrt{\alpha_t}\Lat_{t-1}, (1-\alpha_t) I\right), \\
    p(\Lat_T) & = \Normal\left(\Lat_T | 0, I\right), \\
    p_\theta(\Lat_{t-1}|\widetilde{\Mat}_t) & = \Normal\left(\Lat_{t-1}|\mu_{\theta}(\widetilde{\Mat}_t,t), \tilde{\beta}^2_tI\right),
\end{align*}
where $\alpha_t\in(0,1)$ controls signal retention in the forward process and $\tilde{\beta}_t^2$ is the reverse variance.
For a stochastic reverse update, the transition log-probability is obtained directly from the Gaussian density and summed over the nine lattice entries.
The mean depends on the full crystal state, while the covariance is specified by the diffusion schedule.

\subsection{Fractional Coordinates}
Fractional coordinates live on a periodic domain, so a Wrapped Normal diffusion is used for each atom coordinate:
\begin{align*}
    q(\Frac_t|\Frac_{t-1}) & = \WN\left(\Frac_t|\Frac_{t-1}, (\sigma_t^2 - \sigma_{t-1}^2) I\right), \\
    p(\Frac_T) & = \Unif\left(\Frac_T | 0, 1\right),
\end{align*}
where $\sigma_t \in \mathbb{R}_+$ sets the noise scale at step $t$.
The reverse process is trained with Riemannian score matching \citep{de2022riemannian}, where a network $s_\theta$ approximates the unconditional score used in the sampling procedure of \citet{jiao2023crystal}.

For a sampled fractional coordinate $f'\in[0,1)$, we approximate the wrapped-normal transition log-density by summing over periodic images,
\begin{equation}
    \log p_{\mathrm{WN}}(f'\mid\mu_\theta,s_t)
    \approx \log\sum_{k=-K}^{K}\Normal(f'+k\mid\mu_\theta,s_t^2),
    \label{eq:wrapped-density}
\end{equation}
where $\mu_\theta\in[0,1)$ is the reverse mean modulo one, $s_t$ is the standard deviation of the reverse update, and $K=7$.

\subsection{Atom Types}
Atom types are discrete variables taking values in a fixed vocabulary of $\NumOfTypes$ categories, comprising chemical elements and a vacancy category.
This component is modeled with D3PM \citep{austin2021structured}:
\begin{align*}
    q(\Atom_{t,i}|\Atom_{t-1,i}) & = \Cat\left(\Atom_{t,i}|Q_t\Atom_{t-1,i}^{\text{onehot}}\right), \\
    p(\Atom_{T,i}) & = \Cat\left(\Atom_{T,i} | \one/N_{\text{types}}\right), \\
    p_\theta(\Atom_{t-1,i}|\widetilde{\Mat}_t) & = \Cat\left(\Atom_{t-1,i}|c_{\theta,i}(\widetilde{\Mat}_t,t)\right),
\end{align*}
where $\Atom_{t,i}$ is the type of site $i$, $\Atom_{t-1,i}^{\text{onehot}}$ is its one-hot encoding, and $\one$ is the all-ones vector of length $\NumOfTypes$.
The matrix $Q_t$ mixes probability mass over categories, while $c_{\theta,i}$ gives the reverse categorical probabilities.
For a sampled type $a_i$, the log-probability is simply $\log c_{\theta,i}[a_i]$.
Conditional independence across sites gives the joint log-probability as the sum of these terms, so its evaluation requires no numerical integration or periodic summation.

\section{Related Work}\label{sup:related-works}

Several approaches have been explored for generating stable, unique, and novel (SUN) crystals. We focus on diffusion models trained on known crystal structures, reinforcement learning with stability rewards, and fine-tuning on selected or ranked samples from a generator.

\paragraph{Diffusion models for crystal generation.}
Crystal diffusion models learn the distribution of periodic structures from crystallographic datasets.
MatterGen~\citep{zeni2025generative} jointly diffuses atom types, atomic coordinates, and lattice parameters and supports property-conditioned generation through adapter-based fine-tuning.
DiffCSP~\citep{jiao2023crystal} uses periodic equivariant diffusion over lattices and fractional coordinates for composition-conditioned structure prediction. Its de novo extension also generates atom types.
MiAD~\citep{okhotin2025miad} augments the DiffCSP backbone with mirage atoms.
Transitions between physical and fictitious atoms allow the number of physical atoms to change during denoising.
In this work we use MiAD as a pretrained reference model and fine-tune it using a stability reward. 

\paragraph{Reinforcement learning for generative models.}
Machine-learned interatomic potentials make it practical to estimate crystal energies during fine-tuning. Comparing these energies with a reference convex hull gives a stability signal that can be combined with rewards for novelty and diversity \citep{park2025guiding,prakash2026reinforcement}.
Chemeleon2~\citep{park2025guiding} applies group relative policy optimization (GRPO) to a latent diffusion model. Its frozen decoder maps samples to crystal structures, which are scored using a single-point energy calculation alongside novelty, uniqueness, and diversity criteria.
OMatGRPO~\citep{prakash2026reinforcement} applies GRPO directly to continuous lattice and coordinate dynamics and discrete atom-type transitions in OMatG, which combines stochastic interpolants with discrete flow matching.
Both approaches include MP-20-pretrained generators and report metastable--unique--novel (mSUN) yields.
CrysTune~\citep{govindarajan2026crystune} instead applies GRPO to an autoregressive language model fine-tuned on MP-20 Wyckoff representations.
Its reported trade-off between improved metastability and reduced novelty illustrates why stability alone is insufficient for assessing discovery.

\paragraph{Preference alignment and refinement on generated structures.}
Generated crystals can also supply training data for the next round of fine-tuning. Scores can be used to determine which materials should be retained as supervised examples.
PLaID++~\citep{xu2026plaid} applies iterative direct preference optimization to a language model fine-tuned on MP-20 Wyckoff representations, using generated crystal pairs ranked by stability, novelty, and space-group compatibility.
Crys-JEPA~\citep{liu2026crysjepa} learns an energy-aware crystal representation to screen relaxed generated structures, then refines a diffusion generator through supervised training on selected candidates.
A related baseline is studied by \citet{cao2026reinforcement}: alongside PPO fine-tuning of CrystalFormer, they evaluate supervised fine-tuning on stable generated structures using an Alex-20-pretrained backbone.

Stability thresholds and sample sizes vary across these studies. To compare fine-tuning methods under common conditions, we evaluate DPOK and RTB on the same MiAD backbone at matched generation budgets.
\begin{table*}[t]
    \centering
    \caption{SUN rates (\%) for individual trained checkpoints. Bold: best per method and column.}
\label{tab:ablation_betas}
\centering
\small
\begin{tabular}{c|ccc|ccc}
\toprule
& \multicolumn{3}{c|}{DPOK} & \multicolumn{3}{c}{RTB} \\
\cmidrule(lr){2-4} \cmidrule(lr){5-7}
$\beta$
& SUN@2k & SUN@10k & SUN@100k
& SUN@2k & SUN@10k & SUN@100k \\
\midrule
$0.001$ & 20.40 & 9.80 & 2.61 & 19.55 & 10.47 & 2.71 \\
$0.01$  & \textbf{21.50} & 11.47 & 3.19 & 20.35 & 10.08 & 2.54 \\
$0.1$   & 17.25 & \textbf{15.15} & 9.73 & \textbf{21.80} & \textbf{15.69} & 7.81 \\
$0.5$   & 13.90 & 12.24 & \textbf{9.92} & 17.15 & 15.55 & \textbf{10.32} \\
$2.5$   & 13.65 & 11.42 & 9.87 & 14.80 & 13.47 & 10.31 \\
\bottomrule
\end{tabular}
\par
    \vspace{1.5em}
    \captionsetup{type=table}
    \caption{SUN rates (\%) for pretrained and fine-tuned models. Bold indicates the highest value at each budget.}
\label{tab:final_comparisons}
\centering
\small
\begin{tabular}{lccc}
\toprule
Model & SUN@2k & SUN@10k & SUN@100k \\
\midrule
MiAD & 12.65 & 11.08 & 9.36 \\
Chemeleon2 & 6.65 & 6.99 & 6.19 \\
\midrule
Chemeleon2-GRPO & \textbf{19.05} & 15.05 & 7.70 \\
MiAD-DPOK & 17.25 & 15.15 & 9.73 \\
MiAD-RTB & 17.15 & \textbf{15.55} & \textbf{10.32} \\
\bottomrule
\end{tabular}

\end{table*}
\section{SUN@budget}
\label{sec:sun-budget}

\paragraph{Motivation}
SUN at any finite budget captures the uniqueness of generated structures only at a particular sampling level. 
Higher budgets probe long-term model effectiveness, that is, how well a checkpoint keeps producing distinct, novel, stable crystals as more draws accumulate. 
Lower budgets favor faster collection of a small pack of SUN candidates. 
We take the former goal as more meaningful, because in automated materials search the bottleneck is typically not generative-model inference but the cost of relaxation, stability assessment, and downstream validation.
The standard variant is SUN@10k, and we complement it with SUN@2k and SUN@100k to observe behavior across budgets.

\paragraph{Implementation details}
The generation budget $N$ is the number of crystals drawn from a checkpoint for evaluation. 
We report SUN@$N$ as the percentage of draws that contain at least two elements, are stable, are unique within the sample, and have no structural match in MP-20. 
Stability requires CHGNet energy above the MP convex hull \texttt{2023-02-07-ppd-mp.pkl.gz} from matbench-discovery \citep{riebesell2023matbench} to satisfy $E^{\text{hull}}<0$, so repeated draws of the same structure raise the stable count without adding discoveries. 
Comparisons across checkpoints at a given budget use the same protocol.

\section{Experiments}
\label{sec:experiments}

We evaluate MiAD fine-tuned with DPOK and RTB at 2k, 10k, and 100k draws. Table~\ref{tab:ablation_betas} compares regularization strengths and Table~\ref{tab:final_comparisons} compares the selected models with baselines.

\subsection{Effect of Regularization Strength}

For each method and each $\beta\in\{0.001,0.01,0.1,0.5,2.5\}$, we select the checkpoint with the highest training SUN@2k and report its SUN rate at all three budgets.

The highest SUN@2k occurs at $\beta=0.01$ for DPOK and $\beta=0.1$ for RTB. For both methods, the best-performing setting shifts from $\beta=0.1$ at 10k draws to $\beta=0.5$ at 100k draws.

\subsection{Final Comparisons}

We compare MiAD-DPOK at $\beta=0.1$ and MiAD-RTB at $\beta=0.5$ with pretrained MiAD, pretrained Chemeleon2, and Chemeleon2-GRPO. In Table~\ref{tab:ablation_betas}, these checkpoints have the highest DPOK SUN@10k and RTB SUN@100k, respectively.

Chemeleon2-GRPO has the highest SUN@2k (19.05\%). Both selected MiAD checkpoints outperform Chemeleon2-GRPO at the larger budgets. SUN@10k is 15.15\% for MiAD-DPOK and 15.55\% for MiAD-RTB, compared with 15.05\% for Chemeleon2-GRPO. SUN@100k reaches 9.73\% for MiAD-DPOK and 10.32\% for MiAD-RTB, approximately 26\% and 34\% above the Chemeleon2-GRPO score of 7.70\%.

\section{Conclusion}

\textbf{Both fine-tuning methods outperformed pretrained MiAD and Chemeleon2-GRPO.} Starting from pretrained MiAD, both DPOK and RTB improved SUN rates over the frozen model at every budget tested. Both also outperformed Chemeleon2-GRPO on SUN@10k and SUN@100k. MiAD-DPOK achieved 15.15\% and 9.73\%, respectively, while MiAD-RTB ranked first on both measures at 15.55\% and 10.32\% (Table~\ref{tab:final_comparisons}).

\textbf{Under weak regularization, high early SUN does not persist at larger budgets.} At $\beta=0.01$, the selected DPOK and RTB checkpoints reach SUN@2k of 21.50\% and 20.35\%, but SUN@10k and SUN@100k fall to 11.47\% / 3.19\% and 10.08\% / 2.54\%, respectively (Table~\ref{tab:ablation_betas}). Stronger regularization yields lower SUN@2k and higher rates at the larger budgets.

\textbf{Larger generation budgets place greater demands on the diversity of stable crystals a model can generate.} Among the models in the final comparison, Chemeleon2-GRPO had the highest SUN rate at 2k draws (19.05\%). Its rate decreased to 15.05\% at 10k draws and 7.70\% at 100k draws. MiAD-DPOK and MiAD-RTB exceeded the Chemeleon2-GRPO rate at both larger budgets, with relative gains of 26.4\% and 34.0\%, respectively, at 100k draws. As the budget grows, each additional discovery must be distinct from a larger set of previously generated structures. A model that repeatedly samples from a less diverse set of stable structures can achieve a high SUN@2k yet yield fewer new discoveries per draw as the draw budget grows.





\clearpage
\bibliography{literature}

@article{okhotin2025miad,
  title={{MiAD}: Mirage Atom Diffusion for De Novo Crystal Generation},
  author={Okhotin, Andrey and Nakhodnov, Maksim and Kazeev, Nikita and Lazarev, Mikhail and Ustyuzhanin, Andrey E. and Vetrov, Dmitry},
  journal={arXiv preprint arXiv:2511.14426},
  year={2025},
  url={https://arxiv.org/abs/2511.14426}
}

@article{park2025guiding,
  title={Guiding Generative Models to Uncover Diverse and Novel Crystals via Reinforcement Learning},
  author={Park, Hyunsoo and Walsh, Aron},
  journal={arXiv preprint arXiv:2511.07158},
  year={2025},
  url={https://arxiv.org/abs/2511.07158}
}

@article{prakash2026reinforcement,
  title={Reinforcement Learning on the Discrete Composition Channel of a Crystal Generator: Validated Gains and Reward Hacking},
  author={Prakash, Pawan and H{\"o}llmer, Philipp and Fuhr, Addis and Hirschfeld, Peter and Ganesh, P. and Martiniani, Stefano and Hennig, Richard},
  journal={arXiv preprint arXiv:2610.03880},
  year={2026},
  url={https://arxiv.org/abs/2610.03880}
}

@inproceedings{govindarajan2026crystune,
  title={{CrysTune}: Crystal Generation via Fine-Tuning of Large Language Models on Wyckoff Representations},
  author={Govindarajan, Prashant and Das, Kishalay and Chandar, Sarath},
  booktitle={ICML AI for Science Workshop},
  year={2026},
  url={https://openreview.net/forum?id=Oe5iihLiiV}
}

@inproceedings{xu2026plaid,
  title={{PLaID++}: A Preference Aligned Language Model for Targeted Inorganic Materials Design},
  author={Xu, Andy and Desai, Rohan and Wang, Larry and Ritz, Ethan T. and Hope, Gabriel},
  booktitle={Proceedings of the 43rd International Conference on Machine Learning},
  series={Proceedings of Machine Learning Research},
  volume={306},
  pages={140903--140922},
  year={2026},
  publisher={PMLR},
  url={https://proceedings.mlr.press/v306/xu26l.html}
}

@article{liu2026crysjepa,
  title={{Crys-JEPA}: Accelerating Crystal Discovery via Embedding Screening and Generative Refinement},
  author={Liu, Nian and Kazeev, Nikita and Dale, Stephen Gregory and Maevskiy, Artem and Zeng, Yuwei and Kubo, Ryoji and Huang, Pengru and Laurent, Thomas and LeCun, Yann and Novoselov, Kostya S. and Bresson, Xavier},
  journal={arXiv preprint arXiv:2605.14759},
  year={2026},
  url={https://arxiv.org/abs/2605.14759}
}

@article{cao2026reinforcement,
  title={Reinforcement Fine-Tuning for Materials Design},
  author={Cao, Zhendong and Wang, Lei},
  journal={Physical Review B},
  volume={113},
  pages={024106},
  year={2026},
  doi={10.1103/45zh-44bg},
  url={https://arxiv.org/abs/2504.02367}
}

@inproceedings{fan2023dpok,
  title={{DPOK}: Reinforcement Learning for Fine-tuning Text-to-Image Diffusion Models},
  author={Fan, Ying and Watkins, Olivia and Du, Yuqing and Liu, Hao and Ryu, Moonkyung and Boutilier, Craig and Abbeel, Pieter and Ghavamzadeh, Mohammad and Lee, Kangwook and Lee, Kimin},
  booktitle={Advances in Neural Information Processing Systems},
  volume={36},
  pages={79858--79885},
  year={2023},
  url={https://arxiv.org/abs/2305.16381}
}

@inproceedings{venkatraman2024amortizing,
  title={Amortizing intractable inference in diffusion models for vision, language, and control},
  author={Venkatraman, Siddarth and Jain, Moksh and Scimeca, Luca and Kim, Minsu and Sendera, Marcin and Hasan, Mohsin and Rowe, Luke and Mittal, Sarthak and Lemos, Pablo and Bengio, Emmanuel and Adam, Alexandre and Rector-Brooks, Jarrid and Bengio, Yoshua and Berseth, Glen and Whitammer, Esmeralda S.},
  booktitle={Advances in Neural Information Processing Systems},
  volume={37},
  pages={76080--76114},
  year={2024},
  url={https://arxiv.org/abs/2405.20971}
}

@article{ho2020denoising,
  title={Denoising diffusion probabilistic models},
  author={Ho, Jonathan and Jain, Ajay and Abbeel, Pieter},
  journal={Advances in neural information processing systems},
  volume={33},
  pages={6840--6851},
  year={2020}
}

@article{de2022riemannian,
  title={Riemannian score-based generative modelling},
  author={De Bortoli, Valentin and Mathieu, Emile and Hutchinson, Michael and Thornton, James and Teh, Yee Whye and Doucet, Arnaud},
  journal={Advances in neural information processing systems},
  volume={35},
  pages={2406--2422},
  year={2022}
}

@article{austin2021structured,
  title={Structured denoising diffusion models in discrete state-spaces},
  author={Austin, Jacob and Johnson, Daniel D and Ho, Jonathan and Tarlow, Daniel and Van Den Berg, Rianne},
  journal={Advances in neural information processing systems},
  volume={34},
  pages={17981--17993},
  year={2021}
}

@article{jiao2023crystal,
  title={Crystal structure prediction by joint equivariant diffusion},
  author={Jiao, Rui and Huang, Wenbing and Lin, Peijia and Han, Jiaqi and Chen, Pin and Lu, Yutong and Liu, Yang},
  journal={Advances in Neural Information Processing Systems},
  volume={36},
  pages={17464--17497},
  year={2023}
}

@article{zeni2025generative,
  title={A generative model for inorganic materials design},
  author={Zeni, Claudio and Pinsler, Robert and Z{\"u}gner, Daniel and Fowler, Andrew and Horton, Matthew and Fu, Xiang and Wang, Zilong and Shysheya, Aliaksandra and Crabb{\'e}, Jonathan and Ueda, Shoko and others},
  journal={Nature},
  volume={639},
  number={8055},
  pages={624--632},
  year={2025},
  publisher={Nature Publishing Group UK London}
}

@article{deng2023chgnet,
  author = {Deng, Bowen and Zhong, Peichen and Jun, KyuJung and Riebesell, Janosh and Han, Kevin and Bartel, Christopher J. and Ceder, Gerbrand},
  title = {{CHGNet} as a pretrained universal neural network potential for charge-informed atomistic modelling},
  journal = {Nature Machine Intelligence},
  volume = {5},
  pages = {1031--1041},
  year = {2023},
  doi = {10.1038/s42256-023-00716-3}
}

@article{barroso2024omat24,
  title={{Open Materials 2024 (OMat24)} Inorganic Materials Dataset and Models},
  author={Barroso-Luque, Luis and Shuaibi, Muhammed and Fu, Xiang and Wood, Brandon M. and Dzamba, Misko and Gao, Meng and Rizvi, Ammar and Zitnick, C. Lawrence and Ulissi, Zachary W.},
  journal={arXiv preprint arXiv:2410.12771},
  year={2024},
  url={https://arxiv.org/abs/2410.12771}
}

@article{schmidt2024alexandria,
  title={Improving machine-learning models in materials science through large datasets},
  author={Schmidt, Jonathan and Cerqueira, Tiago F. T. and Romero, Aldo H. and Loew, Antoine and J{\"a}ger, Fabian and Wang, Hai-Chen and Botti, Silvana and Marques, Miguel A. L.},
  journal={Materials Today Physics},
  volume={48},
  pages={101560},
  year={2024},
  doi={10.1016/j.mtphys.2024.101560}
}

@article{riebesell2023matbench,
  title={{Matbench Discovery}--A framework to evaluate machine learning crystal stability predictions},
  author={Riebesell, Janosh and Goodall, Rhys EA and Benner, Philipp and Chiang, Yuan and Deng, Bowen and Lee, Alpha A and Jain, Anubhav and Persson, Kristin A},
  journal={arXiv preprint arXiv:2308.14920},
  year={2023}
}
\bibliographystyle{icml2027}

\newpage
\appendix
\onecolumn
\section{Experimental details}\label{sup:exp-details}

\begin{table}[!ht]
    \centering
    \caption{Fine-tuning settings for the MiAD checkpoints in Table~\ref{tab:final_comparisons}.}
    \label{tab:training-settings}
    \small
    \begin{tabular}{@{}lcc@{}}
        \toprule
        Setting & MiAD-DPOK & MiAD-RTB \\
        \midrule
        Pretrained checkpoint & \multicolumn{2}{c}{MiAD on MP-20, epoch 8000} \\
        Components fine-tuned & \multicolumn{2}{c}{Lattice, fractional coordinates, atom types} \\
        Denoising steps & \multicolumn{2}{c}{100} \\
        Effective batch size & \multicolumn{2}{c}{96} \\
        Optimizer & \multicolumn{2}{c}{Adam} \\
        Model learning rate & \multicolumn{2}{c}{$10^{-4}$} \\
        Gradient norm clipping threshold & \multicolumn{2}{c}{8} \\
        Checkpoint selection & \multicolumn{2}{c}{Highest SUN@2k, evaluated every 10 epochs} \\
        \midrule
        Learning rate for $\log Z$ & --- & $5\times10^{-3}$ \\
        Regularization strength $\beta$ & 0.1 & 0.5 \\
        Selected epoch & 169 & 189 \\
        \bottomrule
    \end{tabular}
\end{table}

\end{document}